\documentclass[11pt]{article}
\usepackage[]{acl}

\usepackage{times}
\usepackage{latexsym}
\usepackage{enumitem}
\usepackage{subcaption}  
\usepackage[T1]{fontenc}

\usepackage[utf8]{inputenc}

\usepackage{microtype}

\usepackage{inconsolata}
\usepackage{amsmath}
\usepackage{capt-of}

\usepackage{graphicx}
\usepackage{booktabs}
\usepackage{xcolor}
\usepackage{colortbl}
\usepackage{multirow}
\usepackage{makecell}
\usepackage{listings}

\usepackage{listings}
\usepackage{booktabs}

\usepackage[inkscapelatex=false]{svg}

\title{Paying for Too Many Tokens? Valid and Cost-Efficient Multimodal LLM Annotation with Simple Heuristics}

\author{Zhixi Zhu \\
  Johns Hopkins University \\
  \texttt{zzhu86@jhu.edu} \\\And
  Kristina Gligorić \\
Johns Hopkins University \\
  \texttt{gligoric@jhu.edu} \\}

\begin{document}
\maketitle
\begin{abstract}
Vision-Language Models (VLMs) enable video annotation at scale, but costs accumulate quickly: 
processing a typical 60-second short-form video at one frame per second requires millions of tokens. To reduce costs, researchers rely on heuristics such as sampling a subset of frames, compressing videos into image grids, or using only a single modality. However, it remains unclear which heuristics save cost, and whether they preserve the downstream conclusions these annotations enable. To address this gap, we conduct a systematic evaluation of these heuristics using short-form videos, on two computational social science (CSS) tasks: sentiment and topic classification. We evaluate each configuration along three axes the literature typically treats separately: classification accuracy, validity of downstream inference, and per-video token cost. First, we find that accuracy and validity diverge: the highest-accuracy configuration can produce wrong conclusions. Second, 
modality value is not guaranteed: text alone can yield strong performance, indicating that adding modalities can add cost without adding signal. Finally, we find that cost can be decoupled from video length when annotating short-form videos: a single $2\times8$ image grid built via simple shot-transition detection approaches full-video understanding ($\kappa$ within~.05), at $\sim 15\%$ of the token cost. Based on these findings, we derive guidelines that can enable cost-aware VLM annotation in CSS.
\end{abstract}

\section{Introduction}\label{sec:intro}

Key computational social science constructs like 
misinformation \cite{qi2024sniffermultimodallargelanguage}, toxicity 
\cite{kiela2021hatefulmemeschallengedetecting}, humor 
\cite{10.1145/2783258.2783388}, stance 
\cite{10.1145/3140565, liang-etal-2024-multi}, accountability~\cite{saez2026egopolice}, and emotion 
\cite{qi2024multimodalemotionrecognitionvisionlanguage} often manifest in images and videos. The rapid growth of multimodal LLM annotations now enables researchers to study these constructs at scale.

However, annotating multimodal data, especially videos, is expensive and time-consuming, as annotation costs quickly accumulate with the large number of frames that need 
to be processed. To address this gap, researchers rely on heuristics, such as sampling a subset of frames 
\cite{tang2025adaptivekeyframesamplinglong}, compressing videos into 
image grids \cite{kim2024imagegridworthvideo}, or using only a single 
modality.

Two questions remain open. First, \emph{which heuristics save 
cost without sacrificing annotation quality?} Although past work has 
studied some of the individual design choices 
\cite{tang2025adaptivekeyframesamplinglong, hu2025mllmbasedvideoframe}, 
we lack a comprehensive evaluation of how frame sampling, grid 
composition, and modality subsets interact. Second, \emph{could these heuristics result in accurate downstream conclusions?} Prior work on compact video representations 
\cite{kim2024imagegridworthvideo} and keyframe sampling 
\cite{hu2025mllmbasedvideoframe} has focused on video question answering, where the goal is accuracy. However, CSS applications typically \emph{leverage annotations} as inputs to regressions, frequency estimates, and other statistical analyses, where the goal is not high accuracy, but \emph{reaching correct conclusions}. To address this gap, we conduct a systematic evaluation of cost-saving heuristics for VLM-based video annotation. We make the following contributions:

\begin{figure*}[t]
    \centering
    \includegraphics[width=\textwidth]{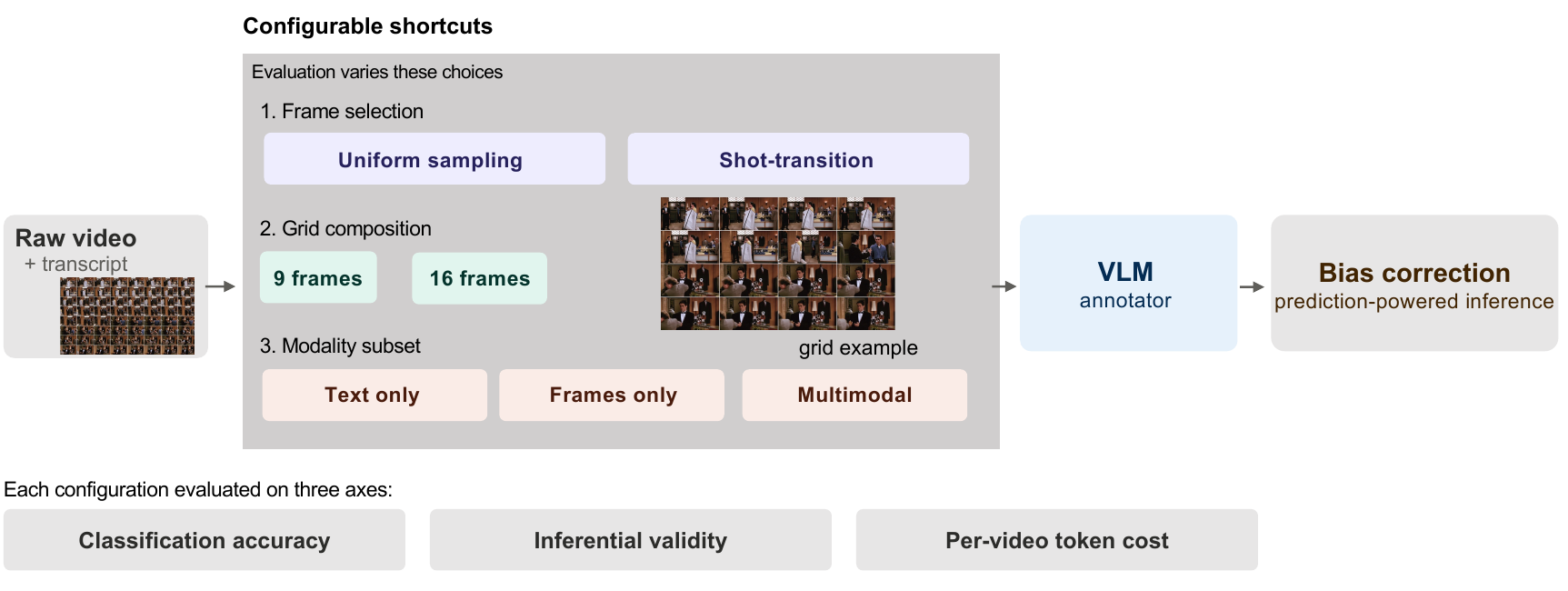}
    \caption{\textbf{Evaluation framework for cost-saving VLM annotation heuristics.} We evaluate cost-saving heuristics for VLM-based video annotation along three configurable choices (frame selection, grid composition, modality), evaluating each configuration on classification accuracy, downstream inferential validity, and per-video token cost. Bias correction via PPI is applied to recover valid confidence intervals from a small set of human labels.}
    \label{fig:comic}
\end{figure*}

\begin{enumerate}[leftmargin=0pt, labelindent=0pt, itemindent=*]

\item \textbf{Benchmarking (Sec.~\ref{methods})}: We evaluate the heuristics used to scale VLM-based video annotation. The evaluation spans sentiment classification on MELD~\cite{poria2019meld}, an 18-way topic classification task on TikTok, and a comparison against full-video understanding. 

\item \textbf{Quantitative insights (Sec.~\ref{results})}: We surface findings not visible from accuracy benchmarks alone: (1) \textbf{annotation and inference accuracy can diverge}: on MELD, effects are missed in multiple configurations, and the highest-accuracy configuration flips the sign of the coefficient relative to ground truth, on TikTok, estimate errors can be large; (2) \textbf{multimodality is task-dependent}: text-only wins on dialogue-driven settings, while multimodal grids win on topic classification where visual and textual cues are complementary; (3) \textbf{compression preserves quality at a fraction of the cost}: a single $2\times 8$ image grid approaches Gemini's full-video understanding ($\kappa$ within .05) at \textbf{14.8\% }of the per-video token cost, at fixed resolution.

\item \textbf{Evidence-based recommendations (Sec.~\ref{discussion})}: We distill these findings into a decision procedure for choosing a compression strategy, deciding when multimodality is worth its cost, and budgeting gold human labels, so that CSS researchers can scale VLM annotation without trading validity for cost. Code, datasets (TikTok video IDs), the shot-transition detector, and the TikTok hashtag query list are released at~\url{https://github.com/JHU-CSS/multimodal-annotation}.
    
\end{enumerate}

\section{Background}

\subsection{VLMs}

Vision-Language Models (VLMs) can jointly understand text and images, and generate text based on the multimodal input~\cite{Ghosh2024ExploringTF}. The development of VLMs has progressed rapidly, with notable models including CLIP \cite{radford2021learningtransferablevisualmodels}, LLaVA \cite{liu2023visualinstructiontuning}, GPT-4V \cite{openai2024gpt4technicalreport}, and Gemini \cite{geminiteam2025geminifamilyhighlycapable}. These models have demonstrated strong capabilities in visual understanding and reasoning, making them increasingly adopted as automated annotators for multimodal data. Most related to our work, \citet{kim2024imagegridworthvideo} use image 
grids as a compact substitute for video input in zero-shot video QA, 
and recent keyframe-sampling work similarly reduces video input to a 
small set of informative frames \cite{hu2025mllmbasedvideoframe}. These 
heuristics are used in practice, but their accuracy has been 
studied primarily on video QA, where the metric of interest is direct 
correctness. We instead evaluate them in a CSS setting, where 
annotations feed into downstream statistical analyses and validity, 
not just accuracy, is the relevant criterion.


\subsection{Inferences based on VLM annotations}

Although VLMs can scale annotation better than human annotators, their predictions likely contain errors. Using these imperfect labels directly in downstream analyses can introduce bias and thus produce invalid estimates. Existing methods such as prediction-powered inference (PPI)~\cite{doi:10.1126/science.adi6000} and design-based supervised learning (DSL)~\cite{10.5555/3666122.3669122} address this by combining a small set of gold-standard human labels with a large set of model predictions to produce valid confidence intervals. The gold-standard labels are used to estimate the bias of the model predictions, without making assumptions about the model's prediction quality. Few studies have explored these repair frameworks with 
multimodal data. Importantly, 
\citet{204946} applied DSL to image analysis 
and showed that even with 90\% accuracy, naive use of AI labels leads 
to poor confidence interval coverage, while DSL achieves nominal 
coverage with only a few hundred human annotations. However, these methods have primarily been applied to static images 
and text. Whether they extend reliably to video data (where 
annotations span hundreds of frames per clip and cost-saving 
heuristics may introduce systematic biases) remains open.

\subsection{Cost in downstream applications}

LLMs are widely used as annotators for text-based tasks like 
sentiment analysis, stance detection, and content moderation, where 
they have been shown to match or even outperform crowd-workers 
\cite{Gilardi_2023}. Beyond labeling, they have also been used 
to generate rationales, instruction-response pairs, and synthetic data 
for training purposes~\cite{wang2023selfinstructaligninglanguagemodels, honovich2022unnaturalinstructionstuninglanguage}. As VLMs became available, researchers started 
applying them to multimodal annotation tasks as well---for example, 
annotating medical images \cite{doi:10.1056/AIoa2300138}. These applications increasingly rely on cost-saving heuristics to make large-scale annotation tractable. While past work has audited token costs~\cite{artola2025your,velascoauditing,sun2025coin}, model choice~\cite{bergemann2025economics,cai2025you}, and compute allocation decisions~\cite{snell2024scaling}, the implications of 
heuristics for downstream inference in the context of multimodal annotation are rarely examined. Our study 
provides the empirical basis for such applications.

\section{Methods}\label{methods}

\subsection{Evaluation setup}

In our study, we evaluate the following configurations: 
the frame-sampling strategies (uniform vs.\ shot-transition), the 
grid composition choices, the modality subsets, and the bias-correction 
step. 
In what follows, we outline the shot-transition + 
image-grid + bias-correction configuration we recommend, since it is \emph{the most general setup} (others are ablations of this setup).

A video typically has a large number of frames. A naive way to perform 
video annotation tasks is to feed all frames together with the prompt 
into the LLM. However, this approach is highly token-inefficient and 
introduces redundancy, since many adjacent frames in videos contain only 
minor visual differences. More importantly, processing every frame may 
not scale well to longer videos, or a larger number of videos, due to the 
context-window length, the computational costs, and inference time, either via API or local inference.

To address these challenges, we reduce the number of frames while 
 preserving essential visual information from the video. The setup we focus on draws inspiration from  
prior work showing that image grids can serve as compact substitutes 
for video input~\cite{kim2024imagegridworthvideo}. Specifically, we 
select a limited set of video frames, i.e., 9 or 16. Then we 
arrange these frames into a composite image grid (see 
Figure~\ref{fig:comic} for an example). This alternative representation 
of a video significantly reduces the number of tokens required, while 
it still preserves information about the temporal progression of the 
video.

A key design choice within this configuration is how to select frames 
from the original video. The simple approach is to sample  
frames at fixed time intervals, uniformly. But uniform sampling might fail 
to capture the most informative moments, especially when there are 
significant visual changes between frames. To address this limitation, 
we employ a heuristic based on shot-transition detection~\cite{smeaton2010video, 10.1145/217279.215266} to sample frames that differ most substantially 
from their neighboring frames. Specifically, for each pair of 
consecutive frames $F_t$ and $F_{t+1}$, we compute a dissimilarity 
score:
\[
D_t = \frac{1}{N} \sum_{i=1}^{N} \left| G_t(i) - G_{t+1}(i) \right| 
      +  \chi^2(H_t, H_{t+1}),
\]
where $G_t(i)$ denotes the grayscale intensity of pixel $i$ and $H_t$ 
denotes the grayscale histogram. Large $D_t$ would indicate major 
visual changes, which may correspond to scene transitions. We identify such transitions 
by selecting local maxima of $\{D_t\}$ above an adaptive threshold, and 
sample frames around these points. The threshold is selected such that the procedure results in the desired number of frames. By doing so, the sampled set of frames is more  likely to capture informative and diverse visual content. More advanced keyframe-selection methods~\cite{tang2025adaptivekeyframesamplinglong} might improve performance further. Here, we consider only simple heuristics that do not introduce additional inference costs to be accounted for.

After constructing the frame representation, we provide it to the VLM 
together with a structured prompt that specifies the annotation task. 
The prompt contains three main components. First, it includes task 
instructions describing what the model should analyze in the video 
frames. The prompt also explains the structure of the visual input, 
indicating that the frames are arranged in a grid and ordered 
chronologically. Second, it defines the valid label space, constraining 
the model to only select from a predefined set of annotation labels. 
Finally, the prompt specifies a fixed output format (JSON), so that the 
model's responses can be parsed  for downstream processing. Full prompting parameters are outlined in the Appendix (Sec.~\ref{app:prompt}).

After collecting annotations from the VLM, we correct downstream statistics while preserving valid inference. For each task, we draw a small labeled correction set uniformly at random from the target dataset and manually annotate it with human labels. This labeled sample is used to estimate the bias of the model-assisted estimator by comparing LLM predictions to human annotations. We then combine this estimated correction with the model predictions on the full dataset to produce bias-corrected estimates and confidence intervals. This is done using prediction-powered inference (PPI) \cite{doi:10.1126/science.adi6000}. The metrics are reported on average, across $n=100$ repeated trials (cf. robustness test with varying number of human annotations in Appendix, Sec.~\ref{app:sensitivity}).

Specifically, for the TikTok hashtag task, the downstream estimands are class frequencies, so we apply PPI to the sample mean of each binary hashtag indicator. For MELD, the downstream estimands are logistic-regression coefficients, so we apply the corresponding PPI estimator for regression. Confidence intervals are constructed by bootstrap resampling of the labeled correction set. We use non-overlapping samples for configuration selection and final inference, i.e., pilot labels used to compare annotation configurations are not reused in the labeled correction set used to report PPI confidence intervals. This procedure lets us leverage scalable LLM-assisted annotation while ensuring that reported estimates reflect uncertainty in the final inferential task rather than overfitting to the model-selection sample. Further details about how PPI was applied are outlined in the Appendix (Sec.~\ref{app:PPI}).

\subsection{Datasets}

We use two datasets: MELD (sentiment, 3 classes) and TikTok (hashtag, 
18 classes). Across both, we evaluate combinations of frame-sampling 
strategy, grid size, and modality subset, with and without PPI correction.

\paragraph{MELD} We sample 900 clips from MELD~\cite{poria2019meld}, 
each a short segment from \textit{Friends} typically containing a dialogue 
between two or more characters; we retain durations between 1 and 6 seconds 
to keep length consistent. Each video in the dataset was annotated by three human annotators, and the ground truth human annotation was obtained by majority voting. Human inter-annotator agreement is high ($\kappa$=0.91)~\cite{poria2019meld}.

The annotation task is to identify a specified 
character's sentiment: the prompt instructs the model to locate the character 
and infer sentiment by jointly considering visual cues (e.g., facial 
expressions) and the transcript, output as structured JSON over a fixed label 
set (positive, negative, neutral).
For the inference task, we study whether predicted sentiments are associated 
with a character's role---main versus supporting cast.

The inference goal is selected because cast type is a salient feature of any character, as it can serve as a fixed covariate for any annotation unit---a setting common in CSS estimation tasks where regressions predict annotation values based on fixed covariates that do not need to be annotated, such as lexical features or author properties~\cite{gligoric2025can,danescu2013computational}. For each sentiment $s$, we fit a binary logistic regression where $Y_i = 1$ indicates that sample $i$ 
expresses $s$ and $X_i \in \{0,1\}$ denotes main-cast membership. Our primary 
statistic is the coefficient $\beta_1$: a positive value indicates main-cast 
characters are more likely to express $s$ than supporting characters, and a 
negative value indicates the opposite. We assess significance via the estimated standard 
error and $p$-value of $\beta_1$.

\paragraph{TikTok} We select eighteen trending TikTok hashtags and collect 
their videos via the TikTok Research API~\cite{tiktok_research_api}, restricting 
to U.S. posts from October 2025--January 2026 with an automatically generated 
transcript (\texttt{voice\_to\_text}), so both modalities are available. Unlike 
MELD, durations vary widely (median $\sim$46s, range $\sim$3s to $\sim$12min). 
The task is to annotate a video's hashtag; the prompt mirrors MELD but restricts 
the label set to these eighteen hashtags. 

We treat hashtags as creator-assigned reference labels: because creators supply 
them, they are used in social media research as signals of a post's topic~\cite{tan2014effect}. Our target is, therefore, not an externally verified content category, but the creator-provided topical label (each video contains a label provided by the creator before model labels are collected). For each hashtag, we estimate its frequency as the proportion of videos assigned to it (a single hashtag can be assigned to a given video). Topic frequency is selected as the objective since CSS researchers often need to monitor such prevalence of content types, across users, or over time~\cite{kirkpatrick2024tiktok,cervi2023tiktok}.

\subsection{Metrics}

We evaluate prediction performance using accuracy and 
Cohen's kappa, which capture overall agreement and agreement beyond 
chance, respectively. Accuracy measures the proportion of correct predictions. We compute both per-class accuracy and the micro-average. Cohen's kappa measures agreement between model predictions and 
ground-truth labels while accounting for expected chance agreements. Costs are computed based on tokenizer estimates.

\begin{table*}[ht!]
\centering

\begin{subtable}{\textwidth}
\centering
\scriptsize
\setlength{\tabcolsep}{3pt}
\begin{tabular}{l r rr rr rrl rr rr}
\toprule
& & \multicolumn{2}{c}{\textbf{\texttt{ShotT} 16 (4$\times$4)}} & \multicolumn{2}{c}{\textbf{Text-only}} & \multicolumn{3}{c}{\textbf{PPI}} & \multicolumn{2}{c}{\textbf{RF}} & \multicolumn{2}{c}{\textbf{RF\,V+T}} \\
\cmidrule(lr){3-4}\cmidrule(lr){5-6}\cmidrule(lr){7-9}\cmidrule(lr){10-11}\cmidrule(lr){12-13}
\textbf{Sent.} & \textbf{GT $\beta$\,[$p$]} & \textbf{$\beta$\,[$p$]} & \textbf{$\Delta$} & \textbf{$\beta$\,[$p$]} & \textbf{$\Delta$} & \textbf{$\beta$\,[$p$]} & \textbf{$\Delta$} & \textbf{95\% CI} & \textbf{$\beta$\,[$p$]} & \textbf{$\Delta$} & \textbf{$\beta$\,[$p$]} & \textbf{$\Delta$} \\
\midrule
Pos. & $.13$\,[$.096$]
     & \cellcolor{green!20}$\mathbf{.05}$\,[$.800$] & $-$.08
     & $.29$\,[$.186$] & $+$.16
     & $.48$\,[$.528$] & $+$.35 & [$-$1.0,~1.9]
     & $-$.61\,[$.039$] & $-$.74
     & $-$.04\,[$.879$] & $-$.17 \\
Neu. & $-$.31\,[$<\!.001$]
     & \cellcolor{orange!20}$.25$\,[$.205$] & $+$.56
     & \cellcolor{orange!20}$.12$\,[$.542$] & $+$.43
     & \cellcolor{green!20}$-\mathbf{.48}$\,[$.411$] & $-$.17 & [$-$1.6,~.7]
     & $-$.82\,[$<\!.001$] & $-$.51
     & $-.58$\,[$.015$] & $-$.27 \\
Neg. & $.27$\,[$<\!.001$]
     & \cellcolor{orange!20}$-$.39\,[$.070$] & $-$.66
     & \cellcolor{red!15}$-$.43\,[$.038$] & $-$.70
     & \cellcolor{green!20}$\mathbf{.19}$\,[$.470$] & $-$.08 & [$-$1.0,~1.5]
     & $2.07$\,[$<\!.001$] & $+$1.80
     & $1.08$\,[$.003$] & $+$.81 \\
\bottomrule
\end{tabular}
\caption{\textbf{MELD: logistic regression estimation accuracy.} Each $\beta$ cell reports the coefficient with its $p$-value in brackets. \colorbox{green!20}{Green} highlights the coefficient closest to ground truth (GT) per sentiment; \colorbox{red!20}{red} a sign-flipped significant coefficient (opposite direction from GT); \colorbox{orange!20}{orange} a missed true effect (GT $p<.05$, method $p\geq.05$). Text-only on Negative is significant in the opposite direction from GT ($\beta{=}{-}.43$, $p{=}.038$ vs.\ GT $\beta{=}{+}.27$, $p{<}.001$).}
\label{tab:meld}
\end{subtable}

\vspace{1em}

\begin{subtable}{\textwidth}
\centering
\scriptsize
\setlength{\tabcolsep}{4pt}
\begin{tabular}{l r rr rr l rr rr}
\toprule
& & \multicolumn{2}{c}{\textbf{\texttt{ShotT} 16 (2$\times$8)}} & \multicolumn{3}{c}{\textbf{PPI}} & \multicolumn{2}{c}{\textbf{RF}} & \multicolumn{2}{c}{\textbf{RF V+T}} \\
\cmidrule(lr){3-4}\cmidrule(lr){5-7}\cmidrule(lr){8-9}\cmidrule(lr){10-11}
\textbf{Hashtag} & \textbf{GT} & \textbf{\%} & \textbf{$\Delta$} & \textbf{\%} & \textbf{$\Delta$} & \textbf{95\% CI} & \textbf{\%} & \textbf{$\Delta$} & \textbf{\%} & \textbf{$\Delta$} \\
\midrule
anime          & 4.4 & 4.0  & $-$0.4 & 6.8  & $+$2.4 & [3.3, 10.2] & 5.4 & $+$1.0 & \cellcolor{green!20}\textbf{4.4} & $+$0.0 \\
basketball     & 5.7 & 5.4  & $-$0.3 & 5.3  & $-$0.4 & [3.8,  6.7] & 5.1 & $-$0.6 & \cellcolor{green!20}\textbf{5.7} & $+$0.0 \\
booktok        & 6.1 & 5.2  & $-$0.9 & \cellcolor{green!20}\textbf{6.9}  & $+$0.8 & [3.9,  9.9] & 7.4 & $+$1.3 & 7.8 & $+$1.7 \\
craft          & 4.3 & 3.9  & $-$0.4 & \cellcolor{green!20}\textbf{4.2}  & $-$0.1 & [1.4,  7.0] & 5.2 & $+$0.9 & 4.7 & $+$0.4 \\
dance          & 5.2 & 4.0  & $-$1.2 & \cellcolor{green!20}\textbf{5.8}  & $+$0.6 & [2.9,  8.8] & {5.8} & $+$0.6 & 6.2 & $+$1.0 \\
dogs           & 6.1 & 6.6  & $+$0.5 & \cellcolor{green!20}\textbf{6.3}  & $+$0.2 & [4.8,  7.9] & 5.8 & $-$0.3 & 5.8 & $-$0.3 \\
food           & 6.6 & 7.7  & $+$1.1 & 8.3  & $+$1.7 & [5.8, 10.9] & \cellcolor{green!20}\textbf{5.7} & $-$0.9 & 5.6 & $-$1.0 \\
funnyvideos    & 5.0 & \cellcolor{red!20}10.7 & $+$5.7 & \cellcolor{green!20}\textbf{7.0}  & $+$2.0 & [3.1, 11.0] & 2.8 & $-$2.2 & 2.2 & $-$2.8 \\
gaming         & 5.2 & 6.6  & $+$1.4 & \cellcolor{green!20}\textbf{5.2}  & $+$0.0 & [1.7,  8.7] & 6.9 & $+$1.7 & 7.1 & $+$1.9 \\
gym            & 6.6 & 5.8  & $-$0.8 & 5.9  & $-$0.7 & [3.0,  8.8] & \cellcolor{green!20}\textbf{6.1} & $-$0.5 & 5.3 & $-$1.3 \\
homedecor      & 5.7 & 5.0  & $-$0.7 & 3.9  & $-$1.8 & [1.8,  5.9] & \cellcolor{green!20}\textbf{6.0} & $+$0.3 & 6.4 & $+$0.7 \\
makeup         & 5.3 & 4.7  & $-$0.6 & 3.6  & $-$1.7 & [1.6,  5.6] & 5.8 & $+$0.5 & \cellcolor{green!20}\textbf{5.4} & $+$0.1 \\
music          & 5.3 & 4.9  & $-$0.4 & 3.9  & $-$1.4 & [0.1,  7.7] & \textbf{5.1} & $-$0.2 & 4.4 & $-$0.9 \\
skincare       & 4.7 & 5.6  & $+$0.9 & \cellcolor{green!20}\textbf{4.6}  & $-$0.1 & [2.4,  6.8] & 5.7 & $+$1.0 & 6.2 & $+$1.5 \\
strangerthings & 4.8 & 4.4  & $-$0.4 & 5.3  & $+$0.5 & [3.0,  7.6] & 4.3 & $-$0.5 & \cellcolor{green!20}\textbf{4.7} & $-$0.1 \\
tech           & 6.2 & 5.7  & $-$0.5 & 4.3  & $-$1.9 & [1.3,  7.2] & 5.1 & $-$1.1 & \cellcolor{green!20}\textbf{6.3} & $+$0.1 \\
travel         & 6.0 & 4.0  & $-$2.0 & \cellcolor{green!20}\textbf{6.8}  & $+$0.8 & [3.3, 10.2] & 4.3 & $-$1.7 & 5.0 & $-$1.0 \\
wealth         & 6.8 & 6.0  & $-$0.8 & 6.0  & $-$0.8 & [3.0,  8.9] & 7.4 & $+$0.6 & \cellcolor{green!20}\textbf{6.7} & $-$0.1 \\
\bottomrule
\end{tabular}
\caption{\textbf{TikTok: per-hashtag frequency estimation accuracy.} Frequency (\%) and deviation from GT ($\Delta$). PPI uses \texttt{ShotT}~2$\times$8 LLM predictions as the proxy with bias correction ($n{=}100$ labeled, $N{=}900$ unlabeled, 95\% CI). \colorbox{green!20}{Green} highlights the configuration whose frequency is closest to GT, per topic. Raw predictions 
distort the TikTok frequencies (e.g., \textit{funnyvideos}, $+5.7\%$ error, highlighted in \colorbox{red!20}{red}); bias 
correction reduces these errors.}
\label{tab:freq}
\end{subtable}

\caption{\textbf{High classification accuracy does not guarantee valid downstream 
inference, but bias correction helps recover it.} Downstream estimation accuracy on 
(a) MELD  and (b) TikTok. RF = random forest on Vertex AI video embeddings; RF V+T = random forest on Vertex AI video + transcript embeddings} 
\label{tab:downstream}
\end{table*}

\subsection{Baselines}

We additionally consider two naive sets of baselines: a traditional machine learning 
approach and LLM-based approaches using unimodal information. For the machine learning baseline, we train a Random Forest classifier on input features that are embeddings obtained via Google Vertex AI's multimodal embedding model. There are two variants: a random forest based on video embeddings alone (RF), and a random forest based on embeddings of the video and its text transcript (RF V+T).

We also include LLM baselines where only one modality of information 
is provided. Although all LLMs are prompted with similar text 
instructions, the actual content about the video is restricted to a 
single modality. In particular, we consider two settings: (1) text-only, 
where the model receives transcripts, and (2) frames-only, where the 
model receives video frames. This allows us to evaluate how much each 
modality contributes on its own.

\subsection{Models}

For all LLM-based approaches, we use GPT-4o-mini as the main model, and additionally test other  models supporting multimodal inputs: Claude Sonnet~4.6, GPT-5.1, 
and Grok~4.3. The temperature is 
set to 0.3 to allow for limited randomness while maintaining 
stable outputs. The prompt  has six components: problem 
contextualization, input format, critical instructions, task description, 
allowed label sets, and JSON output format. A full example of the 
prompt used for MELD annotation is provided in the Appendix~\ref{app:prompt}.

\section{Results}\label{results}

We present results along the three axes--classification accuracy, downstream 
inferential validity, and per-video token cost---and organize them 
around the three findings: accuracy and validity diverge 
(Section~\ref{sec:acc-validity}), modality value is task-dependent 
(Section~\ref{sec:modality}), and cost can be decoupled from video 
length (Section~\ref{sec:cost}). For each dataset, we compare uniform 
frame sampling (\texttt{Uni}) and shot-transition sampling (\texttt{ShotT}) at multiple 
grid sizes, alongside text-only, frames-only, Random Forest, and full-video baselines. To limit the number of model–configuration combinations, we report results with GPT-4o-mini throughout; 
we observe similar trends with other tested models (Claude Sonnet~4.6, GPT-5.1, and Grok~4.3), which we defer to Appendix~\ref{app:models} for conciseness.

\subsection{Accuracy and validity diverge}
\label{sec:acc-validity}

On MELD (Appendix Table~\ref{tab:accuracy}), Text-only achieves the highest 
overall accuracy (.610, $\kappa{=}.398$), followed by the multimodal 
configurations: \texttt{ShotT} 4$\times$4 (.589, $\kappa{=}.371$) ranks highest 
among them, with \texttt{Uni} 4$\times$4 (.571) and \texttt{ShotT} 3$\times$3 (.567) close 
behind. Frames-only trails substantially (.442). On TikTok 
(Appendix Table~\ref{tab:accuracy}), the ordering reverses: all four multimodal 
configurations (.819--.834, $\kappa$ up to .825) outperform every 
baseline, with \texttt{Uni} 3$\times$3 highest. Text-only (.713) and 
Frames-only (.712) are roughly tied and far below the multimodal 
configurations.~\footnote{Note the difference in grid orientation: The 4$\times$4 grid on MELD and the 2$\times$8
grid on TikTok both contain 16 frames and incur the same per-image token cost — they differ only in aspect ratio, chosen to match each dataset's native video orientation (horizontal clips, vertical short-form TikTok); the frame budget is matched.} Random Forest baselines (.507--.560 on MELD; 
.717--.718 on TikTok) underperform LLM-based annotation on TikTok, and trail the strongest MELD configurations, indicating that \textbf{LLM reasoning extracts more from the same inputs than 
traditional feature-based approaches}.

The same configurations look very different along the validity axis 
(Table~\ref{tab:meld}). On MELD, PPI achieves valid coverage of the 
ground-truth estimates for all three sentiments. In contrast, every baseline leads to 
misleading inference. Both LLM baselines miss the true 
Neutral effect; the multimodal LLM (\texttt{ShotT} 4$\times$4) additionally 
misses Negative; and \emph{the Text-only baseline---the 
highest-accuracy configuration---produces a sign error on Negative} 
($\beta{=}{-}.43$, $p{=}.038$ vs.\ GT $\beta{=}{+}.27$, $p{<}.001$), 
in which a researcher would conclude an effect in the wrong direction. 
RF falsely detects significance on Positive ($\beta{=}{-}.61$, 
$p{=}.039$) when the ground-truth effect is not significant 
($p{=}.096$), and distorts the Negative coefficient 
($\Delta{=}{+}1.80$) even when the sign is preserved.

\begin{table*}[t!]
\small
\centering
\begin{tabular}{lrrrr}
\toprule
\textbf{Method} & \textbf{Avg.\ Tokens} & \textbf{Avg.\ Cost (USD)} & \textbf{Min Tokens} & \textbf{Max Tokens} \\
\midrule
3$\times$3 grid (9 frames)           &    36,596 & \$0.00549 &  25,501 &     48,169 \\
2$\times$8 grid (16 frames)          &    47,665 & \$0.00715 &  25,501 &     48,169 \\
1 FPS ($\sim$ 60 frames)   & 2,275,411 & \$0.34131 & 110,505 & 25,821,335 \\
\bottomrule
\end{tabular}
\caption{\textbf{Grid sampling decouples token cost from video length.} 
Average GPT-4o-mini token cost per video by frame input method on the 
TikTok dataset (n=900, avg.\ duration 62.3s). Packing frames into a single 
grid costs $\sim$37--48k tokens regardless of video length, whereas 1~FPS 
sampling scales with duration and is $\sim$50$\times$ more expensive on 
average (and over 500$\times$ in the worst case). The two grids share the 
same max-token value ($\sim$48k), reflecting the API's internal per-image 
token cap, which bounds cost regardless of grid resolution.}
\label{tab:token_methods}
\end{table*}

On TikTok (Table~\ref{tab:freq}, Figure~\ref{fig:ppi}), the raw LLM 
(\texttt{ShotT} 2$\times$8) over-predicts funnyvideos by 5.7 percentage points 
(10.7\% vs.\ GT 5.0\%) and under-predicts travel by 2.0 points (4.0\% 
vs.\ GT 6.0\%). PPI mitigates these biases: all 18 GT class 
frequencies fall within the 95\% confidence intervals, and the 
funnyvideos overestimation drops from 5.7 to 2.0 percentage points. 
PPI produces wide intervals where LLM bias is large---funnyvideos 
$[3.1, 11.0]$, anime $[3.3, 10.2]$---which reflects the limits of 
correction with a small labeled sample, but signals (rather than 
hides) the underlying bias.

In summary, we find that the accuracy ranking and the validity ranking disagree. \textbf{ Accuracy is not a sufficient quality criterion when the annotations feed into downstream inference, and bias correction is required.
}

\subsection{Modality value is task-dependent}
\label{sec:modality}

The two datasets disagree on whether multimodal input helps (Appendix Table~\ref{tab:accuracy}). On MELD, 
Text-only (.610, $\kappa{=}.398$) outperforms every multimodal 
configuration; \texttt{ShotT} 4$\times$4 (.589) beats Frames-only (.442) by a 
wide margin, indicating that adding the transcript is what closes the 
gap---the frames themselves carry little additional signal 
beyond what the dialogue conveys. On TikTok, the opposite holds: 
Text-only (.713) and Frames-only (.712) perform similarly, while 
\emph{any} multimodal grid (.819--.834) substantially exceeds both. 
Visual and textual cues are genuinely complementary for hashtag 
classification. The pattern is robust to other configuration choices. On MELD, all 
multimodal LLM variants except \texttt{Uni} 9 (3$\times$3) outperform RF (.507) and RF V+T (.560) 
regardless of sampling strategy or grid layout. On TikTok, all four multimodal 
LLM variants (.819--.834) outperform Frames-only (.712), Text-only 
(.713), RF (.717), and RF V+T (.718) across every combination of 
sampling strategy and grid size.

Overall, we find that \textbf{multimodality is not universally beneficial}. On 
dialogue-driven sentiment classification, the transcript already 
carries the construct, and adding frames pays a cost without adding 
signal. On hashtag classification, the visual channel carries 
information that the transcript cannot supply. 

\subsection{Cost can be decoupled from video length}
\label{sec:cost}

The image-grid representation produces near-constant per-video cost 
regardless of video duration, since a single composite image is 
subject to the VLM API's per-image token cap. On TikTok (average 
duration 62.3s), a $3{\times}3$ grid averages 36,596 tokens per video 
and a $2{\times}8$ grid averages 47,665, while standard 1-FPS sampling 
averages 2,275,411 tokens, amounting to a $\sim$50$\times$ difference that grows 
linearly with video length. For a 60-second 
TikTok, the grid representation costs $\sim$\$0.007 per video at 
GPT-4o-mini pricing versus $\sim$\$0.34 for full-frame processing, at the same resolution.

We find that cost saving does not come at proportional loss of quality. On a shared
subset of 100 videos, Gemini uses an average of 2,332 tokens per video for a single
\texttt{ShotT} 2$\times$8 grid, compared with 15,774 tokens per video for
full-video comprehension, or $14.8\%$ as many tokens. Holding the model fixed, a single \texttt{ShotT} 2$\times$8 grid with Gemini matches full-video understanding in aggregate accuracy and agreement ($.870$, $\kappa{=}.860$; Appendix Table~\ref{tab:gemini}). For comparison, the same grid representation with GPT-4o-mini achieves $.820$ ($\kappa{=}.806$). These results suggest that, when salient information is preserved, substantial compression of the video input does not necessarily degrade annotation quality.

The compression strategy depends on shot density. Within the grid-based regime, the best sampling strategy depends on 
how much temporal variation a video contains. On MELD (clips 1--6s 
with few shot transitions), shot-transition and uniform sampling show no statistically significant difference at the same grid size (\texttt{ShotT} 4$\times$4: .589 vs.\ \texttt{Uni} 4$\times$4: .571; diff = +0.018, 95\% CI [-0.007, +0.042], p = 0.177;  Table~\ref{tab:bootstrap_sampling}). On TikTok (videos averaging 62.3s with many shots), uniform sampling matches or slightly beats shot-transition 
sampling (\texttt{Uni} 3$\times$3: .834 vs.\ \texttt{ShotT} 3$\times$3: .827)---once 
uniform sampling covers enough shots, targeting transitions adds 
little. Figure~\ref{fig:acc vs duration} shows that TikTok annotation 
accuracy is lowest for very short videos.

The key takeaway is that \emph{cost decouples from video length}. \textbf{A single image grid  approaches full-video understanding} at $\sim$14.8\% of the token cost, 
and the cost is bounded by the VLM API's per-image cap, regardless of 
length. Shot-transition sampling helps when 
the frame budget is tight; uniform sampling 
suffices otherwise.

\section{Discussion}\label{discussion}

\paragraph{Accuracy benchmarks underdetermine annotation quality.}
The missed effects, sign-flips, or frequency overestimations are examples of a problem that is not 
specific to the studied datasets. Any pipeline that improves accuracy by exploiting 
features uncorrelated with the construct of interest can move the estimate in the wrong direction. Text-only annotation 
captures sentiment via linguistic cues that correlate with 
\textit{cast role} (main characters get more emotionally varied 
dialogue), and that correlation is what produces the wrong-sign 
coefficient. The accuracy metric does not account for this because it 
averages over examples. The implication for practice is that an accuracy-only model selection procedure can lead to inaccurate downstream inference.

\paragraph{When the construct lives in one modality, the other 
modalities only add cost.} The MELD/TikTok contrast suggests a simple 
heuristic for whether multimodality is worth its cost: estimate 
\textit{unimodal sufficiency}---the accuracy of the best 
single-modality annotation relative to the multimodal version---on 
a small pilot. On MELD, Text-only exceeds the best 
multimodal accuracy (.610 vs.\ .589); on TikTok, 
Text-only reaches only 85\% of the multimodal kappa (.696 vs.\ 
.825) and Frames-only essentially the same (.695). The construct in 
TikTok hashtags requires both channels; the construct in 
\textit{Friends} dialogue does not. We expect the heuristic to 
generalize: tasks where one channel carries the construct (dialogue 
sentiment, written sentiment over images of text, OCR-able 
documents) reward dropping the other; tasks where construct cues 
are distributed (memes, multimodal stance, 
content-driven hashtags) reward combining them.

\paragraph{Why cost saving is possible.}
The cost reduction we report (Section~\ref{sec:cost}) follows from two effects: the token cap and duration scaling. Within a single model, a grid representation pays one cap regardless of grid resolution or video length, while a per-frame representation pays the cap once per frame. The implication is that for studies of videos longer than a few seconds, the cost difference between the two regimes grows linearly with duration without any change in quality regime, until the video length exceeds what fits intelligibly in a single grid. Relative to Gemini's native full-video understanding specifically: Gemini's video tokenization scales
linearly with duration ($\sim$281 tokens/second in our data), while the \texttt{ShotT}~2$\times$8 grid's token count is
duration-independent. The two cross over at roughly 9 seconds of video; below this threshold the grid is not cheaper, but savings grow with
duration, reaching $\sim$97\% for our longest clips 
against the $\sim$14.8\% average we report across this dataset's videos.
Within the durations we tested, i.e., up to twelve minutes
(Figure~\ref{fig:acc vs duration}), there is no degradation at the 
longest videos; the upper bound of this regime should be investigated in future work.

\paragraph{PPI correction fixes inferential validity, not point accuracy.} We note that the PPI 
correction is a population-level quantity, so the statistical guarantee is for the aggregate estimand rather than for any individual prediction (consistent with~\citeauthor{krsteski2026valid}). Corrected frequencies are therefore not always closer to the reference than the uncorrected ones (homedecor moves from $-0.7$ to $-1.8$), and precision is uneven for the same reason: classes where model bias is large carry more uncertainty in the estimated correction and end up with wider intervals (funnyvideos $[3.1, 11.0]$ vs.\ basketball $[3.8, 6.7]$). In practice, researchers should report the interval and read its width as a signal of how much bias the correction had to account for.

\subsection{Practical guidelines}
\label{sec:procedure}

We summarize the  findings as a procedure that researchers can apply when adopting VLM-based annotation for a new CSS task. The procedure is derived from two short-form video tasks, so the specific values we recommend should be 
re-estimated for tasks with different label spaces, durations, or genres.

\noindent\textbf{1) Estimate unimodal sufficiency.}
Annotate a pilot of $\sim$50 clips per modality. If a single modality 
reaches a high fraction of the multimodal $\kappa$, drop the other---reducing the cost with negligible quality loss. If both trail substantially, 
proceed multimodal.

\noindent\textbf{2) Choose frame selection by shot density.}
The direction we observed was that shot-transition sampling helped on short clips with few transitions, while uniform sampling was sufficient once videos contained many shots. Differences between sampling strategies were small in our data, so we suggest comparing both on the pilot.

\noindent\textbf{3) Choose grid size.}
Default to the smallest grid whose pilot $\kappa$ is acceptable. As shown in Appendix~\ref{app:analyses}, a 3$\times$3 grid can suffice, exceeding the ``almost perfect agreement'' threshold, at a fraction of the cost of full-video understanding.

\noindent\textbf{4) Budget for gold labels and apply statistical correction.}
The required number of gold labels depends on the
estimand: in our sensitivity analysis (Appendix~\ref{app:sensitivity}),
class-frequency estimates are stable from $n_h \approx 100$,
whereas regression coefficients require $n_h \geq 200$, the
smallest budget at which the coefficient error is small in our
data. Apply PPI to all downstream 
estimates, and report PPI 95\% CIs. Treat wide intervals as a 
diagnostic: they indicate either large
VLM bias on that class or an insufficient correction sample. Importantly, defaults such as ``use multimodal,'' or ``trust accuracy'' are not safe without a pilot.

\section*{Limitations}

First, our work covers two datasets of moderate scale (900 videos each) with relatively 
constrained label spaces. Whether the findings generalize to larger 
corpora, longer videos, or finer-grained label sets remains open---in 
particular, the cost-decoupling result holds up to the durations we 
tested (median 46s, max 701s on TikTok), but very long videos may 
exceed what fits intelligibly in a single grid. We study single-label annotation (a single hashtag or a single sentiment can apply). Future work should extend to multi-label annotation.

Second, we treat dataset-provided labels as ground truth, but these labels are 
themselves noisy: MELD sentiment labels reflect annotator judgment on 
sometimes-ambiguous clips, and TikTok hashtags are user-assigned and 
capture creator intent (including reach-seeking behavior) rather than 
objective content. This affects both axes of our evaluation: accuracy 
estimates are bounded by reference quality, and PPI's validity 
guarantee is \emph{relative to} the chosen reference. On TikTok, this means the accuracy metric measures how well a configuration reproduces the creator's assignment, and PPI recovers valid inference about the creators' hashtag-assignment behavior, 
not necessarily about content. Correction operates relative to whatever reference it is given and does not convert creator labels into verified content categories. Future work pairing the creator-assigned 
labels with content-grounded gold labels would help.

Third, the accuracy/validity divergence is a property of using imperfect labels in downstream inference and should hold for any VLM whose errors correlate with 
the regression covariates. The cost-decoupling result depends on the 
VLM API's per-image token cap, which is currently standard but could 
change. We therefore note that the 14.8\% cost result is contingent on specific models, pricing assumptions, and API image-token policies. The task-dependent value of multimodality may not hold across all datasets and models: the recommendation is to test empirically on a pilot, not to default to text-only.

Finally, our study can be extended beyond PPI to include methods such as DSL~\cite{10.5555/3666122.3669122}, or adaptive sampling of human labels~\cite{gligoric2025can}. Similarly, we tested a simple shot transition heuristic. Further cost reductions may be obtained with more advanced keyframe-selection methods~\cite{tang2025adaptivekeyframesamplinglong,hu2025mllmbasedvideoframe}, but with a potential overhead of the inference costs for keyframe selection itself that should be taken into account.

\section*{Ethical considerations}

This work studies heuristics that reduce the cost of VLM-based video annotation, and we note several risks that accompany making such annotation cheaper and more scalable. First, lowering the cost of large-scale multimodal annotation also lowers the barrier to applications we do not endorse, such as surveillance, profiling, or monitoring of individuals and communities. The same pipeline that estimates aggregate topic prevalence could be repurposed to track specific users. We intend our methods for aggregate, population-level analysis in computational social science, not individual-level inference. Second, we treat dataset-provided labels as references, but these encode subjective and at times strategic human judgments rather than objective content categories. Conclusions therefore reflect annotator and creator behavior, and should not be read as ground truth about content, especially where they might inform consequential decisions. Third, while prediction-powered inference corrects aggregate bias, it does not guarantee uniform error rates across subgroups: a VLM may systematically misannotate content from underrepresented groups, dialects, or non-English speakers, and our evaluation does not disaggregate performance along these lines. Researchers applying these heuristics to demographically sensitive constructs should audit subgroup error and treat the bias-corrected intervals as a floor, not a guarantee.

Finally, our TikTok data is collected through the official Research API in compliance with its terms and contains only publicly available videos. We release only video IDs and query lists rather than raw media, and we encourage downstream users to respect platform terms. The TikTok API provides researchers with a limited and non-exclusive license to study public user and content data. The MELD dataset is available under GPL-3.0 license. The analyses are consistent with the intended uses.

\bibliography{custom}

\appendix

\section{Appendix}
\label{sec:appendix}

\begin{table*}[t]
\centering
\scriptsize
\setlength{\tabcolsep}{3.5pt}
\begin{tabular}{l cccccc cc}
\toprule
& \multicolumn{6}{c}{\textbf{VLM-based (GPT-4o-mini)}} & \multicolumn{2}{c}{\textbf{Random Forest}} \\

\cmidrule(lr){2-7} \cmidrule(lr){8-9}
\textbf{MELD: Sentiment} & \textbf{Frames-only} & \textbf{Text-only} & \textbf{\texttt{ShotT} 16 (4$\times$4)} & \textbf{\texttt{ShotT} 9 (3$\times$3)} & \textbf{\texttt{Uni} 16 (4$\times$4)} & \textbf{\texttt{Uni} 9 (3$\times$3)} & \textbf{RF} & \textbf{RF V+T} \\
\midrule
Negative  & .309 & .572 & .516 & \textbf{.583} & .575 & .581&  .361 & .288 \\
Neutral   & .514 & .628 & .592 & .502 & .507 & .470 &.770 & \textbf{.874} \\
Positive  & .482 & .627 & .689 & .684 & \textbf{.705} & .586 & .145 & .275 \\
\midrule
\textbf{Overall}  & .442 & \cellcolor{green!20}\textbf{.610} & .589 & .567 & .571& .530 & .507 & .560 \\
\textbf{$\kappa$} & .139 & \cellcolor{green!20}\textbf{.398} & .371 & .350 & .360& .365 & .157 & .250 \\

\cmidrule(lr){2-7} \cmidrule(lr){8-9}
\textbf{TikTok: Hashtag} & \textbf{Frames-only} & \textbf{Text-only} & \textbf{\texttt{ShotT} 16 (2$\times$8)} & \textbf{\texttt{ShotT} 9 (3$\times$3)} & \textbf{\texttt{Uni} 16 (2$\times$8)} & \textbf{\texttt{Uni} 9 (3$\times$3)} & \textbf{RF} & \textbf{RF V+T} \\
\midrule
anime          & .800 & .625 & .825 & .850 & .825 & .850 & \textbf{.875} & .850 \\
basketball     & .863 & .706 & .902 & .902 & \textbf{.922} & .902 & .843 & .902 \\
booktok        & .782 & .800 & .836 & \textbf{.891} & .855 & .873 & .764 & .800 \\
craft          & .641 & .795 & .821 & \textbf{.846} & \textbf{.846} & .821 & .744 & .692 \\
dance          & .660 & .319 & .638 & .702 & .723 & .702 & .745 & \textbf{.787} \\
dogs           & .891 & .691 & \textbf{.982} & \textbf{.982} & .964 & \textbf{.982} & .855 & .855 \\
food           & .746 & \textbf{.932} & .881 & .898 & .881 & .915 & .644 & .610 \\
funnyvideos    & \textbf{.711} & .667 & \textbf{.711} & .689 & .689 & .689 & .200 & .156 \\
gaming         & .894 & .596 & .872 & .894 & .872 & \textbf{.915} & .894 & .872 \\
gym            & .712 & .644 & .831 & .831 & .831 & \textbf{.847} & .780 & .729 \\
homedecor      & .588 & .784 & .863 & \textbf{.882} & .843 & .863 & .843 & .843 \\
makeup         & \textbf{.812} & .667 & .792 & .792 & .771 & \textbf{.812} & .750 & .750 \\
music          & .521 & .708 & \textbf{.729} & .708 & \textbf{.729} & .708 & .583 & .562 \\
skincare       & .857 & .857 & \textbf{.929} & \textbf{.929} & .905 & .905 & .881 & \textbf{.929} \\
strangerthings & .744 & .628 & .860 & .884 & \textbf{.907} & \textbf{.907} & .674 & .698 \\
tech           & .571 & \textbf{.804} & .786 & .750 & \textbf{.804} & .786 & .571 & .607 \\
travel         & .426 & .685 & .630 & .630 & .685 & \textbf{.722} & .519 & .593 \\
wealth         & .656 & \textbf{.836} & \textbf{.836} & .820 & .787 & .803 & .754 & .705 \\
\midrule
\textbf{Overall}   & .712 & .713 & .819 & .827 & .824 & \cellcolor{green!20}\textbf{.834} & .717 & .718 \\
\textbf{$\kappa$}  & .695 & .696 & .808 & .816 & .814 & \cellcolor{green!20}\textbf{.825} & .700 & .701 \\
\bottomrule
\end{tabular}

\caption{\textbf{The value of the visual modality is task-dependent.} {Per-class accuracy across annotation configurations on MELD (sentiment, 3 classes) and TikTok (hashtag, 18 classes).} On MELD, Text-only achieves the highest overall accuracy ($\kappa$=0.398), while on TikTok all multimodal grid configurations substantially outperform both unimodal baselines ($\kappa$ 0.808--0.825 vs.\ 0.695--0.696), illustrating that modality value is task-dependent. \colorbox{green!20}{Green} highlights the best-performing configuration overall.}
\label{tab:accuracy}
\end{table*}

\subsection{Prompting}
\label{app:prompt}

The following prompt was used for emotion and sentiment analysis of video frames on the MELD dataset. The prompt is provided along with a grid of frames sampled chronologically from each clip, as well as the corresponding transcript.

\begin{lstlisting}[float=false]
You are a video analyst. You are given a <K>x<L> grid of frames from the TV show *Friends*, sampled in chronological order from a short video clip. Analyze the facial expressions in these frames and the tone of the transcript together to infer emotion.
The frames are arranged left to right, top to bottom, and separated by white lines.

# INPUT
You will receive:
- The name of one specific character (<CharacterName>) to analyze.
- A grid of <N> frames (visual input).

# CRITICAL INSTRUCTION
- You MUST use the exact character name provided in the prompt (<CharacterName>).
- Do NOT modify, abbreviate, replace, or omit it.
- Do NOT use any other character's name.
- Always include this exact name as the JSON key in your output.

# TASK
1. Determine whether the specified character (<CharacterName>) is **visible** in the <N>-frame grid.
   - If clearly present in any of the frames (face or body visible), mark as visible.
   - If completely absent, mark as not visible.

2. If the character **is visible**, infer their **emotion** and **sentiment** using the image.

3. If the character **is not visible**, set both emotion and sentiment labels to "N/A" and confidences to null.

Allowed emotion labels: ["joy", "sadness", "fear", "surprise", "disgust", "angry", "neutral"]
Allowed sentiment labels: ["positive", "negative", "neutral"]

# OUTPUT FORMAT
Return **only valid JSON**, following this exact structure:

{
  "<CharacterName>": {
    "character_visible": <true or false>,
    "emotion": {
      "label": "<one of the 7 labels or 'N/A'>",
      "confidence": <integer 1-10 or null>
    },
    "sentiment": {
      "label": "<one of the 3 labels or 'N/A'>",
      "confidence": <integer 1-10 or null>
    }
  }
}
\end{lstlisting}

The following prompt was used for hashtag classification of videos on the TikTok dataset. The prompt is provided along with a grid of frames sampled chronologically from each clip, along with the transcript.
\begin{lstlisting} [float=false]
You are a TikTok short-form video classifier. Given a <KxL> frame grid (<N> frames sampled in chronological order from a short vertical video, laid out left-to-right across two rows) and a transcript, choose exactly one hashtag from the provided list (DO NOT invent new hashtags). 
Favor visuals over transcript if they conflict.
Respond ONLY with JSON: {"hashtag": "...", "confidence": 1-10, "rationale": "..."}

Hashtag options:
- anime
- basketball
- booktok
- craft
- dance
- dogs
- food
- funnyvideos
- gaming
- gym
- homedecor
- makeup
- music
- skincare
- strangerthings
- tech
- travel
- wealth

Transcript:
<transcript>
\end{lstlisting}

\begin{figure}[t]
\centering
\includegraphics[width=\linewidth]{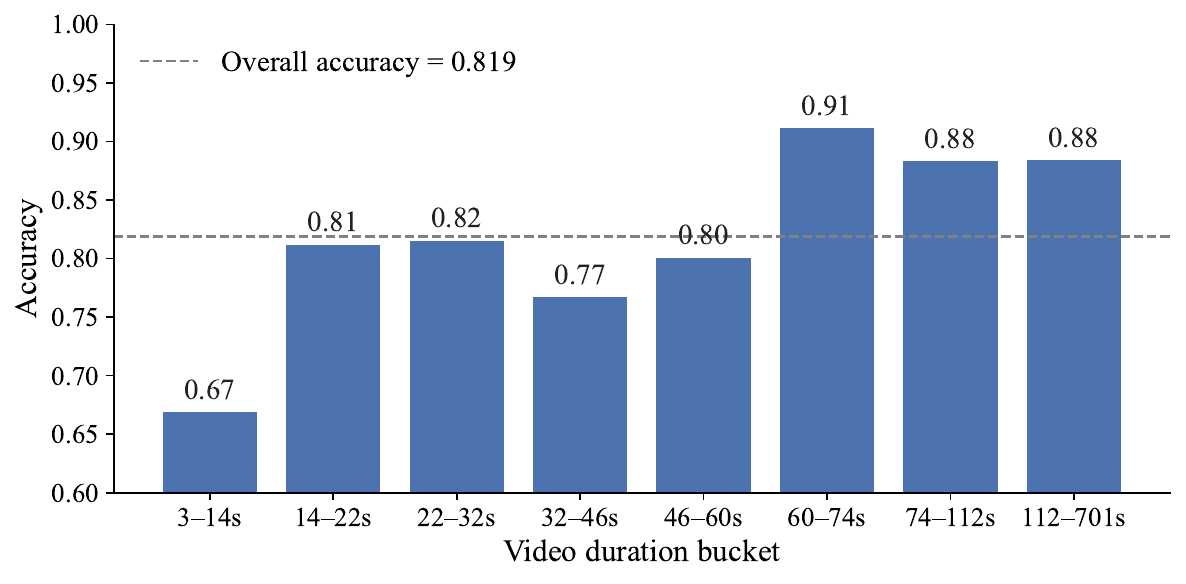}
\caption{Accuracy of TikTok video annotations by video duration (Shot Transition Detection $2 \times 8$).}
\label{fig:acc vs duration}
\end{figure}
\subsection{PPI details}\label{app:PPI}

\begin{figure}[t]
\centering
\includegraphics[width=\linewidth]{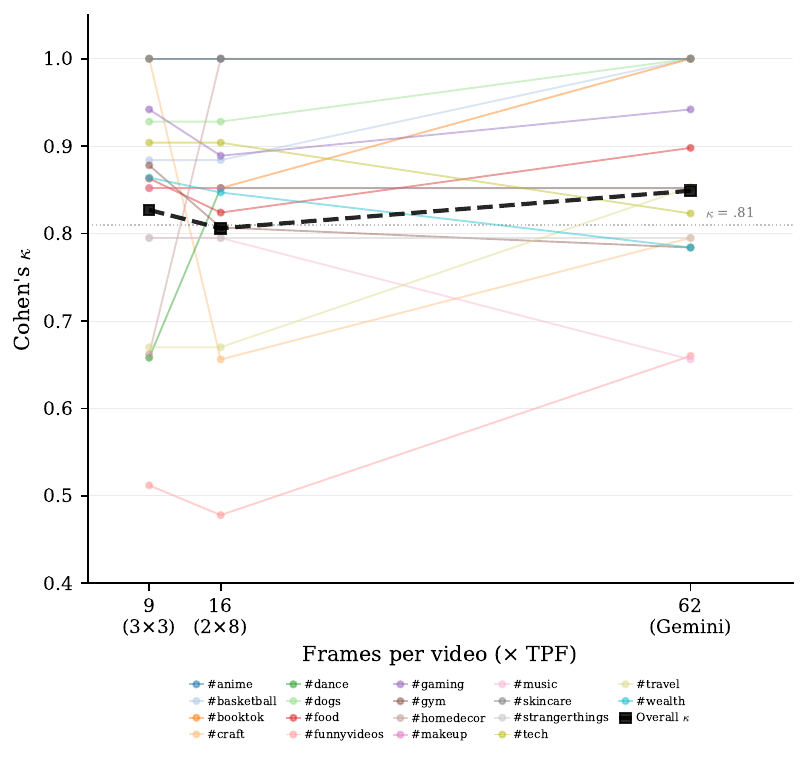}
\caption{Annotation cost vs.\ quality on TikTok. The x-axis shows frames per video (in units of tokens-per-frame, TPF): 9 ($3 \times 3$ grid), 16 ($2 \times 8$ grid), and 62 (Gemini at 1 FPS, since TikTok videos average 62.3s). The dashed line at $\kappa = .80$ marks the conventional ``almost perfect agreement'' threshold \citep{landis1977measurement}. The \texttt{ShotT} $3 \times 3$ grid method exceeds this threshold at a fraction of the per-video token cost.}
\label{fig:cost}
\end{figure}

For each downstream estimand, we use non-overlapping labeled samples for configuration selection and final inference. We first draw a pilot set of 100 videos uniformly at random from the target dataset and use this set only to compare annotation configurations, including modality subsets, frame-sampling strategies, and grid sizes. After selecting the final configuration, we draw a separate correction set of 100 videos uniformly at random from the remaining data. Human annotations on this correction set are used only for prediction-powered inference and are not reused for model or configuration selection. For TikTok, where the estimands are class frequencies, we apply the standard PPI mean estimator separately to each binary hashtag indicator. For MELD, where the estimands are logistic-regression coefficients, we use the PPI estimating-equation formulation: we compute the logistic-regression score using model-predicted labels on the full dataset and correct this score using the average difference between the score evaluated with human labels and model labels on the correction set. Confidence intervals are obtained by bootstrap resampling the correction set while holding the full set of model predictions fixed. The reported intervals, therefore, quantify uncertainty from the finite human correction sample for the observed dataset. Results are reported on average, across $n=100$ repeated trials.

\subsection{Additional analyses}\label{app:analyses}

\begin{table}[ht!]
\centering
\small
\begin{tabular}{l cc}
\toprule
\textbf{Method} & \textbf{Accuracy} & \textbf{$\kappa$} \\
\midrule
Gemini (video)       & \textbf{.870} & \textbf{.860} \\
\texttt{ShotT} 2$\times$8 (Gemini)      & \textbf{.870} & \textbf{.860} \\
\texttt{ShotT} 2$\times$8 (GPT-4o-mini) & .820 & .806 \\
\bottomrule
\end{tabular}
\caption{Accuracy and Cohen's $\kappa$ on 100 common videos ($N{=}100$).}
\label{tab:gemini}
\end{table}

\begin{table*}[t]
  \centering
  \small
  \setlength{\tabcolsep}{5pt}
  \caption{Accuracy and Cohen's $\kappa$ across modality and frame-sampling conditions, with bootstrap CIs}
  \label{tab:bootstrap_sampling}
  \begin{tabular}{llcc}
    \toprule
    Dataset & Method & Accuracy [95\% CI] & Cohen's $\kappa$ [95\% CI] \\
    \midrule
    \multirow{4}{*}{MELD}
      & \textbf{Frames-only} & 0.442\ [0.410, 0.474] & 0.139\ [0.090, 0.187] \\
      & \textbf{Text-only}   & 0.610\ [0.579, 0.641] & 0.398\ [0.351, 0.446] \\
      & \textbf{\texttt{ShotT} 16} (4$\times$4) & 0.589\ [0.557, 0.621] & 0.371\ [0.323, 0.419] \\
      & \textbf{\texttt{Uni} 16}  (4$\times$4)   & 0.571\ [0.539, 0.603] & 0.360\ [0.313, 0.406] \\
    \addlinespace
    \midrule
    \multirow{4}{*}{TikTok ($n{=}900$)}
      & \textbf{Frames-only} & 0.712\ [0.683, 0.742] & 0.696\ [0.665, 0.727] \\
      & \textbf{Text-only}   & 0.713\ [0.683, 0.743] & 0.696\ [0.664, 0.728] \\
      & \textbf{\texttt{ShotT} 9} (3$\times$3) & 0.827\ [0.802, 0.851] & 0.816\ [0.790, 0.842] \\
      & \textbf{\texttt{Uni} 9} (3$\times$3)   & 0.834\ [0.810, 0.859] & 0.825\ [0.798, 0.850] \\
    \addlinespace
    \bottomrule
  \end{tabular}
\end{table*}

\begin{figure}[t]
\centering
\includegraphics[width=\linewidth]{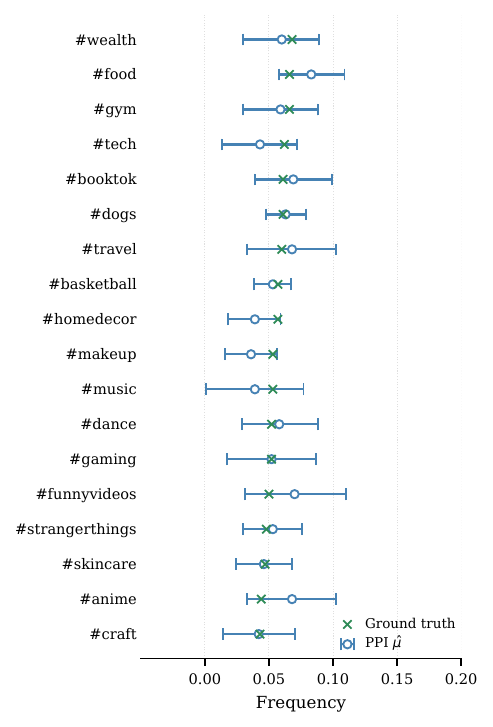}
\caption{Per-hashtag PPI 95\% confidence intervals vs.\ ground truth.}
\label{fig:ppi}
\end{figure}

We show accuracy by video duration in Figure~\ref{fig:acc vs duration} and Per-hashtag PPI 95\% confidence intervals vs.\ ground truth in Figure~\ref{fig:ppi}.

Figure~\ref{fig:cost} plots annotation quality ($\kappa$) against per-video token cost across grid configurations and Gemini full-video understanding. Moving from a $3\times3$ grid (9 frames) to a $2\times8$ grid (16 frames) yields negligible quality gain, while both fall only $\sim$.05 $\kappa$ below Gemini at roughly 14.8\% of its token cost. Notably, the $3\times3$ grid already exceeds the conventional ``almost perfect agreement'' threshold of $\kappa=.80$ \citep{landis1977measurement}, suggesting that further investment in frames or a more expensive model yields diminishing returns for most CSS annotation tasks.


\begin{table}[t]
  \centering
  \footnotesize
  \setlength{\tabcolsep}{3pt}
  \renewcommand{\arraystretch}{1.1}
  \caption{MELD sentiment ($N{=}100$). Each cell: accuracy / Cohen's $\kappa$. \textbf{Bold} = best per model.}
  \label{tab:meld_perclass}
  \resizebox{\columnwidth}{!}{%
  \begin{tabular}{l c c c c c}
    \toprule
    Model & Text-only & \makecell{Img-only\\(4$\times$4 s.)} & \makecell{Img+Text\\(3$\times$3 s.)} & \makecell{Img+Text\\(4$\times$4 s.)} & \makecell{Img+Text\\(4$\times$4 u.)} \\
    \midrule
    Claude Sonnet 4.6 & .61 / .39 & .52 / .18 & .59 / .37 & \textbf{.61 / .41} & .60 / .39 \\
    GPT-5.1           & \textbf{.60 / .37} & .48 / .16 & .42 / .17 & .48 / .24 & .51 / .29 \\
    Grok 4.3          & \textbf{.64 / .40} & .41 / .09 & .40 / .16 & .59 / .33 & .62 / .39 \\
    GPT-4o-mini       & .56 / .32 & .48 / .13 & .56 / .33 & \textbf{.60 / .38} & .60 / .37 \\
    \bottomrule
\end{tabular}}

\end{table}

\begin{table}[ht!]
  \centering
  \footnotesize
  \setlength{\tabcolsep}{3pt}
  \renewcommand{\arraystretch}{1.1}
  \caption{TikTok hashtag ($N{=}100$, 18 classes). Each cell: accuracy / Cohen's $\kappa$. \textbf{Bold} = best per model.}
  \label{tab:tiktok_perclass}
  \resizebox{\columnwidth}{!}{%
  \begin{tabular}{l c c c c c}
    \toprule
    Model & Text-only & \makecell{Img-only\\(2$\times$8 s.)} & \makecell{Img+Text\\(3$\times$3 s.)} & \makecell{Img+Text\\(2$\times$8 s.)} & \makecell{Img+Text\\(2$\times$8 u.)} \\
    \midrule
    Claude Sonnet 4.6 & .78 / .76 & .80 / .78 & \textbf{.88 / .87} & .84 / .83 & .85 / .84 \\
    GPT-5.1           & .77 / .75 & .73 / .71 & \textbf{.87 / .86} & .81 / .79 & .82 / .81 \\
    Grok 4.3          & .75 / .73 & .69 / .67 & \textbf{.87 / .86} & .83 / .82 & .85 / .84 \\
    GPT-4o-mini       & .76 / .74 & .72 / .70 & \textbf{.85 / .84} & .82 / .81 & .84 / .83 \\
    \bottomrule
  \end{tabular}}
\end{table}

\begin{table*}[ht!]
\tiny
\centering
\begin{tabular}{r rccc rccc rccc}
\toprule
 & \multicolumn{4}{c}{Positive} & \multicolumn{4}{c}{Neutral} & \multicolumn{4}{c}{Negative} \\
\cmidrule(lr){2-5}\cmidrule(lr){6-9}\cmidrule(lr){10-13}
$n_h$ & $\beta$ & $\Delta$ & SD & Coverage
     & $\beta$ & $\Delta$ & SD & Coverage
     & $\beta$ & $\Delta$ & SD & Coverage \\
\midrule
 80 & $+$0.50 & $+$0.37 & 3.64 & 91\% & $-$0.30 & $+$0.01 & 0.71 & 96\% & $+$1.23 & $+$0.95 & 3.73 & 91\% \\
100 & $+$0.10 & $-$0.03 & 0.67 & 95\% & $-$0.32 & $-$0.01 & 0.58 & 93\% & $+$0.60 & $+$0.33 & 0.70 & 97\% \\
200 & $+$0.17 & $+$0.04 & 0.51 & 96\% & $-$0.35 & $-$0.04 & 0.36 & 95\% & $+$0.43 & $+$0.16 & 0.41 & 95\% \\
250 & $+$0.12 & $-$0.01 & 0.41 & 98\% & $-$0.38 & $-$0.07 & 0.28 & 99\% & $+$0.46 & $+$0.18 & 0.36 & 98\% \\
450 & $+$0.07 & $-$0.06 & 0.22 & 100\% & $-$0.37 & $-$0.06 & 0.20 & 99\% & $+$0.42 & $+$0.15 & 0.25 & 97\% \\
500 & $+$0.09 & $-$0.04 & 0.23 & 98\% & $-$0.39 & $-$0.08 & 0.17 & 99\% & $+$0.44 & $+$0.16 & 0.20 & 96\% \\
\bottomrule
\end{tabular}
\caption{PPI sensitivity analysis on MELD. For each $n_h$, we sample $n_h$ human-labeled examples from a pool of 1,000 videos, averaged over 100 random draws.
$\Delta = \beta - \beta_{\text{GT}}$.
GT coefficients: Pos.\ $\beta{=}{+}0.13$, Neu.\ $\beta{=}{-}0.31$, Neg.\ $\beta{=}{+}0.27$. Coverage is the fraction of the 100 per-draw 95\% PPI intervals that contain $\beta_{\text{GT}}$. }
\label{tab:meld_sensitivity}
\end{table*}
\subsection{All the models}\label{app:models}

\begin{table*}[h!]

  \centering\tiny\setlength{\tabcolsep}{4pt}
  \renewcommand{\arraystretch}{1.15}

  \begin{tabular}{r cccc cccc cccc}
    \toprule
    & \multicolumn{4}{c}{\#funnyvideos} & \multicolumn{4}{c}{\#travel} & \multicolumn{4}{c}{\#anime} \\
    \cmidrule(lr){2-5}\cmidrule(lr){6-9}\cmidrule(lr){10-13}
    $n_h$ & $\theta$ & $\Delta$ & SD & Cov. & $\theta$ & $\Delta$ & SD & Cov. & $\theta$ & $\Delta$ & SD & Cov. \\
    \midrule
    GT & 5.11\% & --- & --- & --- & 6.01\% & --- & --- & --- & 4.60\% & --- & --- & --- \\
    \midrule
    80  & 4.86\% & $-$0.24 & 2.43 & 89\% & 5.19\% & $-$0.81 & 1.72 & 80\% & 4.14\% & $-$0.46 & 1.51 & 92\% \\
    100 & 5.19\% & $+$0.09 & 1.76 & 95\% & 5.35\% & $-$0.66 & 1.63 & 87\% & 4.16\% & $-$0.45 & 1.25 & 92\% \\
    200 & 5.11\% & $+$0.00 & 1.06 & 99\% & 5.87\% & $-$0.14 & 1.01 & 97\% & 4.43\% & $-$0.17 & 0.80 & 95\% \\
    250 & 4.88\% & $-$0.23 & 0.97 & 96\% & 5.79\% & $-$0.21 & 1.00 & 98\% & 4.43\% & $-$0.18 & 0.74 & 96\% \\
    450 & 5.04\% & $-$0.06 & 0.74 & 99\% & 5.96\% & $-$0.05 & 0.57 & 99\% & 4.54\% & $-$0.07 & 0.46 & 98\% \\
    500 & 4.97\% & $-$0.14 & 0.64 & 100\% & 6.00\% & $-$0.00 & 0.53 & 99\% & 4.52\% & $-$0.08 & 0.42 & 100\% \\
    \bottomrule
  \end{tabular}
  \caption{PPI sensitivity to human-annotation budget ($n_\text{human}$) on TikTok.}
  \label{tab:ppi_sensitivity_tiktok}
\end{table*}

Tables~\ref{tab:meld_perclass} and~\ref{tab:tiktok_perclass} confirm that the 
patterns reported in the main text with GPT-4o-mini are not artifacts 
of a single model. The task-dependence of the visual modality holds across all four 
models: on MELD, text-only inputs are competitive or best for every model---they are 
the top configuration for GPT-5.1 ($.60/.37$) and Grok~4.3 ($.64/.40$), the best 
multimodal setting adds at most four accuracy points (GPT-4o-mini, $.56\!\to\!.60$), 
and image-only inputs collapse uniformly ($\kappa\le.18$ for all models)---whereas on 
TikTok hashtag prediction every model benefits substantially from frames, with 
shot-aligned multimodal inputs raising Cohen's $\kappa$ by 9--13 points over text-only 
(e.g., Grok~4.3, $.73\!\to\!.86$). The cross-model consistency extends to the optimal 
configuration: the sparse $3\times3$ shot grid is best on TikTok for the four models 
on which it is available.

\subsection{Number of human annotations for PPI}\label{app:sensitivity}

To evaluate the impact of the number of human annotations, we conduct a
sensitivity analysis varying the human annotation budget
$n_h \in \{80, 100, 200, 250, 450, 500\}$. We run this analysis on both MELD and TikTok. For each value of $n_h$, we
sample $n_h$ human-labeled examples from a pool of $1,000$ videos---treating
the remaining $1,000 - n_h$ as unlabeled---and average results over $100$
random draws. On MELD, we report the mean PPI coefficient estimate $\beta$ across draws, its deviation
from the ground-truth coefficient ($\Delta = \beta - \beta_{\text{GT}}$), the standard deviation of the point estimates across draws
and empirical coverage
(Table~\ref{tab:meld_sensitivity}). At $n_h=80$, coverage ranges from $91\%$ to $96\%$ across the three
sentiment classes, while the Positive and Negative estimates show
high variability across draws. By $n_h=200$, coverage is at least
$95\%$ for all three classes and across-draw variability is substantially
lower. For Negative, $\Delta$ decreases from $+0.95$ at $n_h=80$ to
$+0.16$ at $n_h=200$, while the SD decreases from $3.73$ to $0.41$. Increasing the budget further does not
 substantially change mean estimates, but it generally reduces variability across draws. 
TikTok shows a similar pattern (Table~\ref{tab:ppi_sensitivity_tiktok}). At $n_h=80$, coverage ranges from $80\%$ to $92\%$, but by $n_h=200$ it reaches at least $95\%$ for all three topics. Variability across draws also decreases as the human annotation budget increases. Overall, the results suggest that a budget between 100 and 200 human annotations is sufficient for reasonably stable estimates, with \(n_h=200\) achieving consistently high coverage across both datasets. Additional human annotations
progressively reduce variability in the estimates.

\end{document}